\documentclass{article}

    \usepackage[final, nonatbib]{neurips_2021}

\usepackage[utf8]{inputenc} %
\usepackage[T1]{fontenc}    %
\usepackage{hyperref}       %
\usepackage{url}            %
\usepackage{booktabs}       %
\usepackage{amsfonts}       %
\usepackage{nicefrac}       %
\usepackage{microtype}      %
\usepackage{xcolor}         %

\usepackage{graphicx}

\usepackage[english]{babel}
\usepackage[backend=biber,
            style=numeric,
            sorting=none,
            maxbibnames=20]
            {biblatex}
\usepackage{csquotes}
\title{Self-supervision of wearable sensors time-series data for influenza detection}

\author{%
  Arinbjörn Kolbeinsson \\
  Evidation Health \\
  \texttt{arinbjorn@evidation.com} \\
   \And
  Piyusha Gade \\
  Evidation Health \\
  \texttt{pgade@evidation.com} \\
   \And
  Raghu Kainkaryam \\
  Evidation Health \\
  \texttt{rkainkaryam@evidation.com} \\
   \And
  Filip Jankovic \\
  Evidation Health \\
  \texttt{fjankovic@evidation.com} \\
   \And
  Luca Foschini \\
  Evidation Health \\
  \texttt{lfoschini@evidation.com} \\
}

\begin{document}

\maketitle

\begin{abstract}
  
  Self-supervision may boost model performance in downstream tasks. However, there is no principled way of selecting the self-supervised objectives that yield the most adaptable models.
  Here, we study this problem on daily time-series data generated from wearable sensors used to detect onset of influenza-like illness (ILI). We first show that using self-supervised learning to predict next-day time-series values allows us to learn rich representations which can be adapted to perform accurate ILI prediction.
  Second, we perform an empirical analysis of three different self-supervised objectives to assess their adaptability to ILI prediction.
  Our results show that predicting the next day's resting heart rate or time-in-bed during sleep provides better representations for ILI prediction.
  These findings add to previous work demonstrating the practical application of self-supervised learning from activity data to improve health predictions.
\end{abstract}

\section{Introduction}

Ground-truth labels for health data are both difficult and expensive to acquire.
Lack of access to labelers (e.g. physicians) and patients not reporting symptoms are common sources of missing labels. 
As a result, many healthcare data are partially or completely unlabelled, with larger prospective studies often having proportionally fewer labels. This trade-off in study size and accuracy is particularly severe in healthcare, although it manifests in most real-world studies across domains.
We are interested in learning from person-generated activity time-series data that has heavily imbalanced or missing labels. As ground-truth labels are a requirement for supervised learning, we must look to other approaches in order to make use of the vast information contained in the unlabelled data.

One potential method to fit this need is transfer learning, where a model is first trained on a different but near-identical population to the final task. However, on closer inspection, this becomes difficult to implement in practice. Firstly, study design critically affects the signals that connect the observed data to the output labels. It is impractical to expect access to a different dataset created through the same study design. Secondly, in the case where we want to study a rare disease, the low prevalence will make it a challenge to find a large, separate initial training set. Therefore, transfer learning alone does not satisfy all our criteria.

Self-supervised learning allows for learning of representations without defined labels.
Although the theoretical principles of these methods have yet to be described, the empirical performance has been widely reported. Most notably in natural language processing \cite{brown2020language} and other syntactic tasks such as programming \cite{chen2021evaluating}, but also in image classification \cite{dosovitskiy2020image}. The taxonomy of these models is debated, with some scholars \cite{bommasani2021foundation} describing pre-trained, self-supervised, meta-learning models as \emph{Foundation Models}. A core concept of self-supervised learning is that the objective is defined as some function of the data itself. However, selecting an optimal self-supervised objective task is not trivial. Here, we study and compare three objective tasks and gauge their performance on influenza-like illness (ILI) prediction with a lab-validated ground-truth.

We take inspiration from natural language processing, where next-word prediction is a task that enables rich representations to be learned \cite{brown2020language}.
Our end-to-end transformer models initially learn through self-supervision to predict next-day activity levels using activity data from the previous ten days in a large population.
These models are then fine-tuned on the ILI prediction task and evaluated on a small dataset with laboratory-confirmed ILI labels. Final evaluation on an ILI prediction task in a held-out testing set reveals that performance is dependent on the self-supervision task.

\section{Data and study set-up}
We use the HomeKit dataset collected by Evidation Health between February and May 2020. \(5\,229\) individuals took part in a prospective cohort study where wearable (Fitbit) device data (including resting heart-rate, step count, and sleep) were recorded. Along with these raw data, Fitbit provides API to multiple processed features that they have pre-calculated, including but not limited to nap count, total minutes asleep, total minutes in bed, total minutes spent sedentary, calories spent on activity and total calories spent.

Every day, each participant was prompted to complete a survey on ILI-related symptoms. If a participant reported two specific symptoms (cough and one of body ache, feeling feverish, chills, sweats) on the same day, they were sent a kit to self-administer a nasal swab which was then tested in a laboratory using an approved polymerase chain reaction (PCR) method. The PCR results revealed that infections identified in the study were caused by a host of respiratory viruses including, but not limited to, influenza virus, respiratory syncytial virus, rhinovirus and SARS‑CoV‑2.

\begin{figure}[!htbp]
    \centering
        \includegraphics[width=0.95\textwidth]{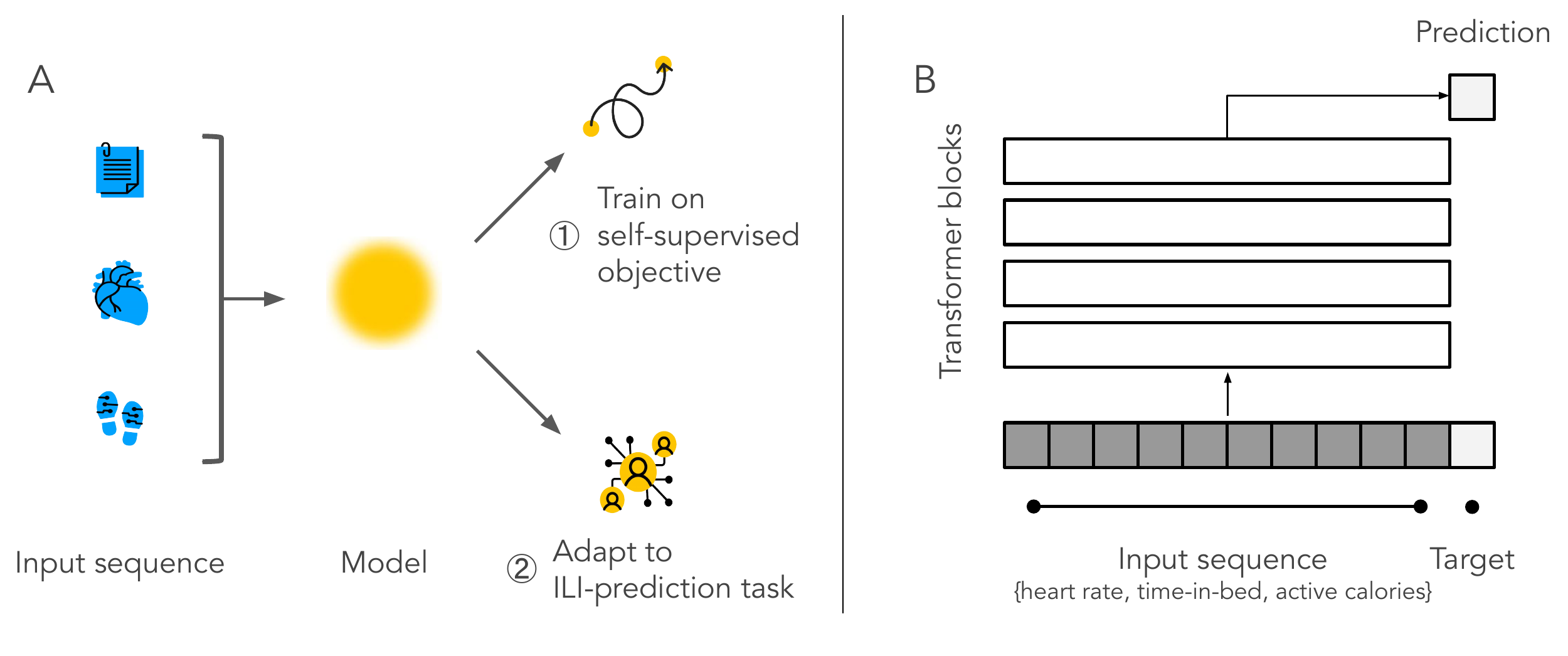}
    \caption{Overview schematic of the study design and model architecture. A) The study design showing how the input data sequences are used to train the model, first on a selected self-supervised objective. The model is then adapted to ILI prediction by retraining only the final layer. B) A simplified architectural overview of the transformer-based model. }
    \label{fig:overview}
\end{figure}

Although label coverage is almost complete in the original study, we devise a set of experiments to simulate a scenario where we have access to two populations: A) a large set of unlabeled activity data and B) a small set of labeled data. In both sets, the recorded activity features (inputs) are identical. A schematic of this is shown in Figure \ref{fig:overview}A.

For self-supervised learning, we train three models to predict next-day resting heart rate, time-in-bed minutes and activity calories expended, respectively. For every model, the input data consisted of resting heart rate, minutes spent in bed, estimated activity calorie output, and missingness flags for each feature.
For evaluation, we apply the model to flu prediction. The input data has the same form as that of SSL training, i.e. ten days of resting heart rate, minutes spent in bed, estimated activity calorie output, and missingness flags for each feature. The label is the laboratory-confirmed cases of respiratory viral infection, not limited to influenza. 

\section{Model and learning}

We use a decoder-only transformer-based architecture, inspired by state-of-the-art natural learning processing approaches \cite{brown2020language}.
It is composed of four transformer blocks. Each block contains a dense self-attention layer with dimensionality \(2\,048\), dropout (\(p = 0.1\)) and ReLU activation. Self attention is calculated as \( attention(Q, K, V) = softmax(QK^{T} / \sqrt{d_k}) V \) \cite{vaswani2017attention}. Where \(Q\), \(K\) and \(V\) are the query, key and value matrices, respectively and \(d_k\) is the dimensionality of the keys.
Layer normalisation was applied after each transformer block. \( LayerNorm(v) = \gamma \frac{v - \mu}{\sigma} + \beta\) \cite{xiong2020layer}. Where \(\mu\) and \(\sigma\) are the mean and standard deviation of the vector, respectively and \(\gamma\) and \(\beta\) are learnable parameters.
Position embeddings for the sequence are learned from random initialization and added to the input sequence. A simplified overview of the model architecture is shown in Figure \ref{fig:overview}B.

The final layer maps the output of the last transformer block to the task-dependent prediction. In regression tasks, this is a fully-connected layer with a single output unit. In the case of binary classification, the output is a vector of length two that, when combined with softmax activation, represent the predicted class pseudo-probabilities.

For self-supervised learning, we train three models for 50 epochs with the objective of minimizing the mean squared error (MSE) of next-day prediction of three features: resting heart rate, time spent in bed, and activity calories expended. The three models have identical architectures as described above and are randomly initialized. In this setting, we only train on sequences where the next-day sequence is known. Fine-tuning on ILI prediction classification was done by minimizing the binary cross entropy of the true and predicted outputs.

All models are minimized with Adam \cite{kingma2014adam}, initial learning rate of \(1\) with cosine annealing and gradient clipping at \(1.0\) to accelerate training \cite{brown2020language, zhang2019gradient}. We train with a batch size of 64 for 50 epochs. For fine-tuning on the adaptation task, we freeze the transformer weights and retrain a new fully-connected output layer from random initialization for 30 epochs. 

\section{Results}

\begin{figure}[!htbp]
    \centering
        \includegraphics[width=0.70\textwidth]{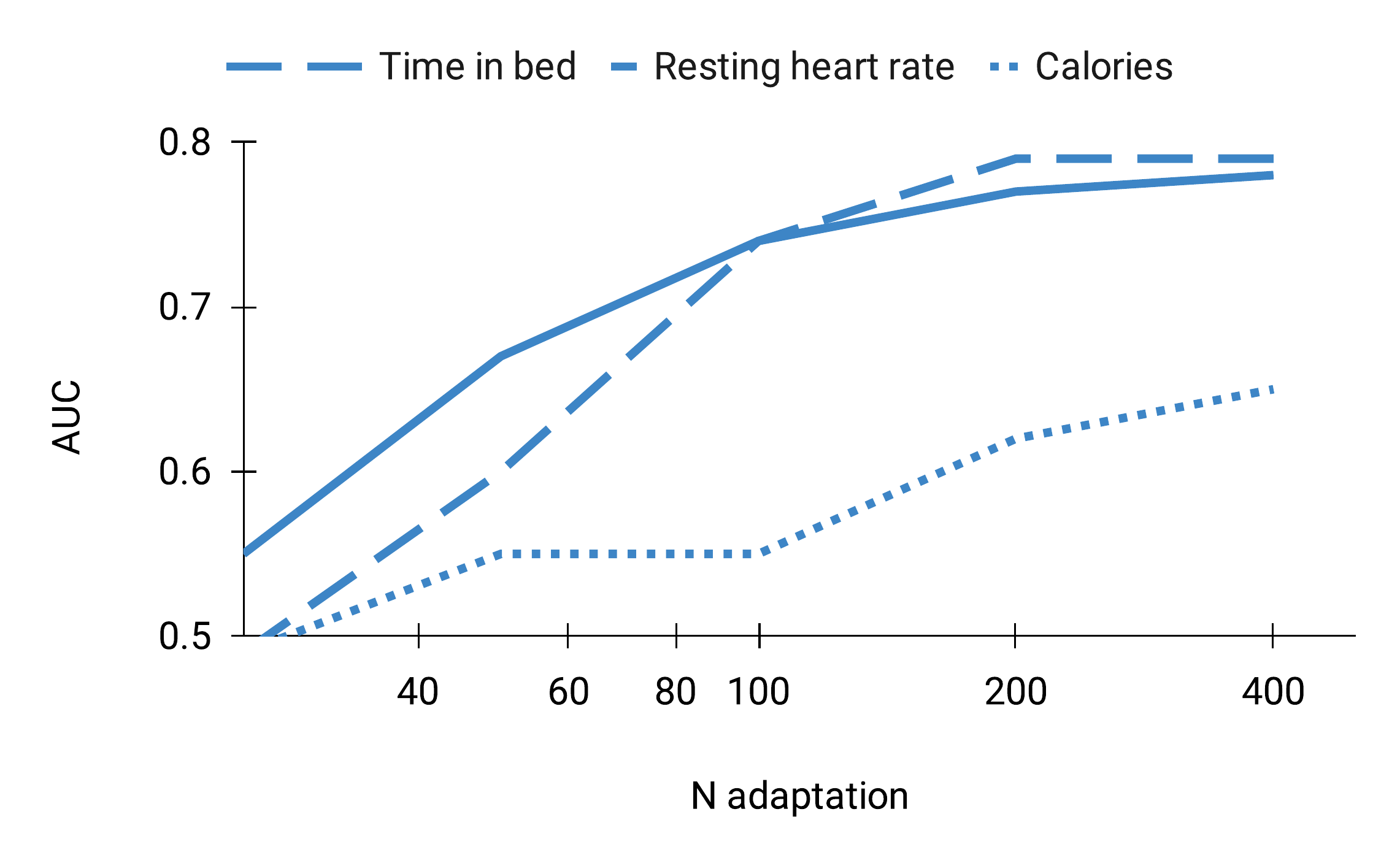}
    \caption{Comparison of models trained with three different self-supervised objective tasks. The AUC (y-axis) quantifies the performance on the adaptation task, ILI prediction. }
    \label{fig:chart}
\end{figure}

We compare adaptation of the three models on five adaptation sets of different sizes, ranging from 25 to 400 individuals. Final evaluation is performed on a held-out test set of 64 individuals. The performance of the three models is visualised in Figure \ref{fig:chart} and listed in Table \ref{tab:chart}. For all three models, the performance, as measured by AUC on the ILI prediction test set, increases logarithmically as the number of samples in the adaptation training step increases.

Above \(\sim200\) adaptation samples, the resting-heart rate and time-in-bed models appear to saturate in performance. However, the results suggest that the model trained on calories might benefit from additional adaptation samples.

Out of the three models, the self-supervised objective on resting-heart rate and time-in-bed gave the strongest result on the downstream task. Training on activity calories expended resulted in a weaker adaptation model.  With only 25 adaptation samples, the resting-heart rate model achieved an AUC of \(0.55\), noticeably better than the other two models. With more adaptation samples, the model adapted from time-in-bed eventually matches the one trained on resting-heart rate.

\begin{table}[!htb]
  \caption{Comparison of ILI prediction performance (area under ROC curve) of models.}
  \label{table}
  \centering
  \begin{tabular}{rccc}
    \toprule
    \multicolumn{1}{c}{}
    &
    \multicolumn{3}{c}{Self-supervised objective}                   \\
    \cmidrule(r){2-4}
    N adaptation    & Resting heart-rate & Time-in-bed & Calories \\
    \midrule
    \(25\)      & \(0.55\)  & \(0.49\)  & \(0.49\)      \\
    \(50\)      & \(0.67\)  & \(0.60\)  & \(0.55\)      \\
    \(100\)     & \(0.74\)  & \(0.74\)  & \(0.55\)      \\
    \(200\)     & \(0.77\)  & \(0.79\)  & \(0.62\)      \\
    \(400\)     & \(0.78\)  & \(0.79\)  & \(0.65\)      \\
    \bottomrule
  \end{tabular}
    \label{tab:chart}
\end{table}

\section{Discussion}
In this work, we showed that self-supervised learning of time-series data provides rich representations that can augment ILI identification. We also explored three different self-supervision objective tasks and found that a model self-trained on predicting next-day resting heart-rate or time-in-bed outperforms a model trained using active calories expended.

The most related work in the literature is a study on transformer-based transfer learning from fatigue prediction to ILI classification \cite{merrill2021transformer}. The significant difference between the two works is that ours uses self-supervision to learn the initial representations instead of transferring from a set of collected labels. Our work also makes use of day-level aggregates versus high-frequency minute-level data. While minute-level features offer significantly more information, they are orders of magnitude more computationally intensive. Other work on wearable sensors and attention-based models is the learning of individualized activity responses for cardiovascular risk-factor prediction \cite{hallgrimsson2018learning}. Self-attention models have also been applied to forecasting influenza prevalence in a population \cite{wu2020deep}.

The application of self-supervised approaches in the field creates its own set of challenges, such as the one we explored here: self-supervision task selection. Fundamentally, this is highly related to the problem of task-relatedness and has been explored in many subfields of machine learning, including transfer learning \cite{pan2009survey} and meta-learning \cite{hospedales2020meta}, and has connections to functional symmetry \cite{lengyel2020genni}.

From a practical perspective, it might be possible to omit task selection and predict all features jointly. Given that heart-rate, sleep and movement are not independent, a model learning all three might capture more diverse relationships and give better results on related downstream tasks, such as ILI prediction. This is an obvious next step for research in this area.

Further improvements can also be made to the study design implemented for the ILI prediction task. Here, we are predicting positive test results, not symptoms. There is a time lag between symptom onset and lab results, on the order of 1 to 3 days. The results must be interpreted with this caveat in mind, as the forecasting capabilities of these models have not been quantified. Additional artifacts for which the effect on the results is unknown include reduced wear-time around illness and the effect of unidentified ILI cases in the self-supervised training set.

\section{Conclusion}
The application of self-supervised learning has great potential in medicine and health, where labels are expensive and often missing.
Our results highlight this potential and demonstrate a practical use case where a large set of unlabelled health data can be used and adapted to a different but related task. 
The remaining open problems offer a large scope for future research which, along with robust baselines and theoretical analysis, would set a strong foundation for applying self-supervised learning for person-generated health data.

\begin{ack}
This project has been funded in whole or in part with Federal funds from the Department of Health and Human Services; Office of the Assistant Secretary for Preparedness and Response; Biomedical Advanced Research and Development Authority, under Contract No. 75A50120C00091.

\end{ack}

\renewcommand{\bibname}{References}
\printbibliography

\end{document}